\documentclass[10pt, a4paper]{article}

\usepackage[]{lrec-coling2024} 

\usepackage{times}
\usepackage{latexsym}
\usepackage{amsmath}
\usepackage{graphicx}
\usepackage{subfigure}
\usepackage[T1]{fontenc}

\usepackage[utf8]{inputenc}

\usepackage{microtype}

\usepackage{inconsolata}
\usepackage{multirow}
\usepackage{booktabs}
\usepackage{tabularray}

\usepackage{booktabs}
\usepackage{vcell}

\usepackage{soul, color, xcolor}

\newcommand{\hsl}[2][]{{\color{black}{#2}}}

\soulregister\cite7 
\soulregister\citep7 
\soulregister\citet7 
\soulregister\ref7 
\soulregister\pageref7 

\newcommand{\BenchmarkName}{LatEval}

\title{\BenchmarkName{}: An Interactive LLMs Evaluation Benchmark with \\ Incomplete Information from Lateral Thinking Puzzles}

\name{Shulin Huang$^{1*}$\thanks{ $^*$ indicates equal contribution. Work is done during Shulin's internship at Netease Games AI Lab.}, Shirong Ma$^{1*}$, Yinghui Li$^{1*}$, Mengzuo Huang$^{2}$, Wuhe Zou$^{2}$,\\ 
{ \bf \large Weidong Zhang$^{2}$ and Hai-Tao Zheng$^{1,3\dagger}$\thanks{ $^{\dagger}$ Corresponding author: Hai-Tao Zheng. (E-mail: zheng.haitao@sz.tsinghua.edu.cn)}
}} 

\address{$^{1}$Shenzhen International Graduate School, Tsinghua University \\ 
        $^{2}$Netease Games AI Lab, HangZhou, China, $^{3}$Peng Cheng Laboratory,  Shenzhen, China \\
        \texttt{\{sl-huang21,masr21,liyinghu20\}@mails.tsinghua.edu.cn}}

\abstract{
With the evolution of LLMs, they are endowed with impressive logical reasoning, or vertical thinking capabilities. But can they think out of the box? Do they possess proficient lateral thinking abilities? Following the setup of Lateral Thinking Puzzles, we propose a novel evaluation benchmark, \BenchmarkName{}, which assesses the model's lateral thinking within an interactive framework. In our benchmark, we challenge LLMs with 2 aspects: 
(1) posing high-quality questions that break out of conventional norms but are beneficial for puzzle-solving. (2) integrating existing information to gradually deduce the truth through reasoning.
We observe that it is hard for most LLMs to accomplish lateral thinking during interactions. 
Even the most powerful LLM, GPT-4, faces challenges in achieving satisfactory performance, and for most open-source models, simply completing this task is quite difficult.
This evaluation benchmark provides LLMs with a highly challenging and differentiating task that is crucial to an effective AI assistant. Our dataset and source codes are available at \url{https://github.com/THUKElab/LatEval}.
 \\ \newline  }

\begin{document}

\maketitleabstract

\section{Introduction}

Large Language Models (LLMs) are gaining increasing capabilities, which enable them to effectively tackle a broad range of tasks~\cite{el2021automatic, hao2022recent, DBLP:journals/corr/abs-2307-09007, cheng2023ml}. 
As LLMs exhibit significant potential, their evaluation has attracted considerable attention and is widely regarded as crucial~\cite{DBLP:journals/corr/abs-2402-11100, DBLP:journals/corr/abs-2311-11268, DBLP:journals/corr/abs-2307-14878}.
Existing mainstream evaluation benchmarks, such as MMLU~\cite{hendrycks2020measuring}, C-Eval~\cite{huang2023c}, GSM8K ~\cite{cobbe2021gsm8k} evaluate models by posing a variety of problems, including problems about mathematics, science, law, and general knowledge, thus reflecting the vertical thinking capacity of models.
\textbf{Vertical thinking, also known as convergent thinking}, refers to a systematic and logical approach that focuses on finding the single best solution to a problem, involving analyzing the problem step-by-step and narrowing down possibilities~\cite{hernandez2008vertical}.

\begin{figure}
    \centering
    \includegraphics[width=0.9\columnwidth]{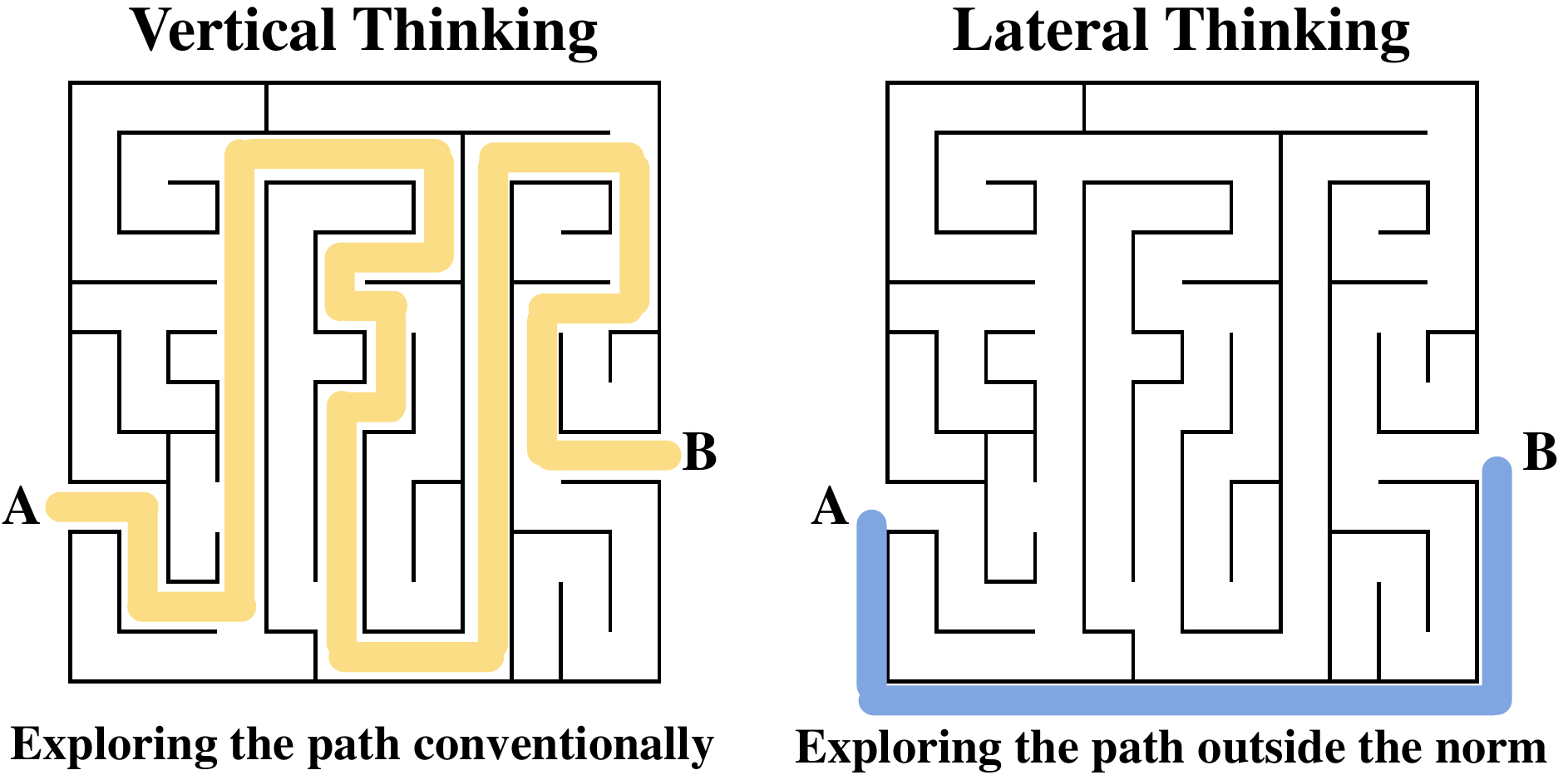}
    \caption{
    The Comparison of Vertical Thinking and Lateral Thinking. 
    Vertical Thinking typically refers to thinking within established or conventional thought patterns, following the known rules. Lateral Thinking involves breaking out of traditional thought patterns and employing innovative approaches to explore non-conventional solutions. 
    }
    \label{fig:intro}
\end{figure}

However, existing benchmarks overlook the model's capability for lateral thinking which also plays a crucial role in human daily thinking. 
\textbf{Lateral thinking, also known as divergent thinking}, is widely recognized in psychology and education. As shown in Figure~\ref{fig:intro}, unlike vertical thinking, lateral thinking involves exploring multiple possibilities, thinking outside the box, and considering unconventional ideas. Lateral thinking encourages individuals to make unexpected connections between seemingly unrelated concepts or ideas, allowing for a broader exploration of potential solutions~\cite{russ1988primary, tsai2012play, russ2013affect}.

Originated from the concept of ``lateral thinking'', Lateral Thinking Puzzles, or namely situation puzzles, are a type of puzzle considered to encourage creative thinking~\cite{ali2019lateral}, which are usually played in a group, with one host and a few players.
A brief story is presented to the players, but it lacks most of the crucial information. The players ask the host a series of ``yes'' or ``no'' questions to gather information, and then piece together the comprehensive truth~\cite{sloane1994great}. The puzzles often involve unexpected twists, hidden clues, or counter-intuitive solutions, which requires players to think beyond traditional patterns and consider unconventional possibilities.

Taking Lateral Thinking Puzzles as the scene, we propose a novel evaluation benchmark, \BenchmarkName{}, which assesses the model's lateral thinking within an interactive framework for the first time.
Our benchmark utilizes the fundamental setup of Lateral Thinking Puzzles, involving one host and one player. The player represents an LLM under evaluation, while a powerful LLM (e.g., GPT-4) serves as the host for automatic evaluation.
The model under evaluation engages in lateral thinking to pose questions based on the given puzzle, interact with the host, and subsequently provide its answer (i.e., deduction) after acquiring sufficient information.

\begin{figure*}[ht]
\centering
\includegraphics[width=0.95\linewidth]{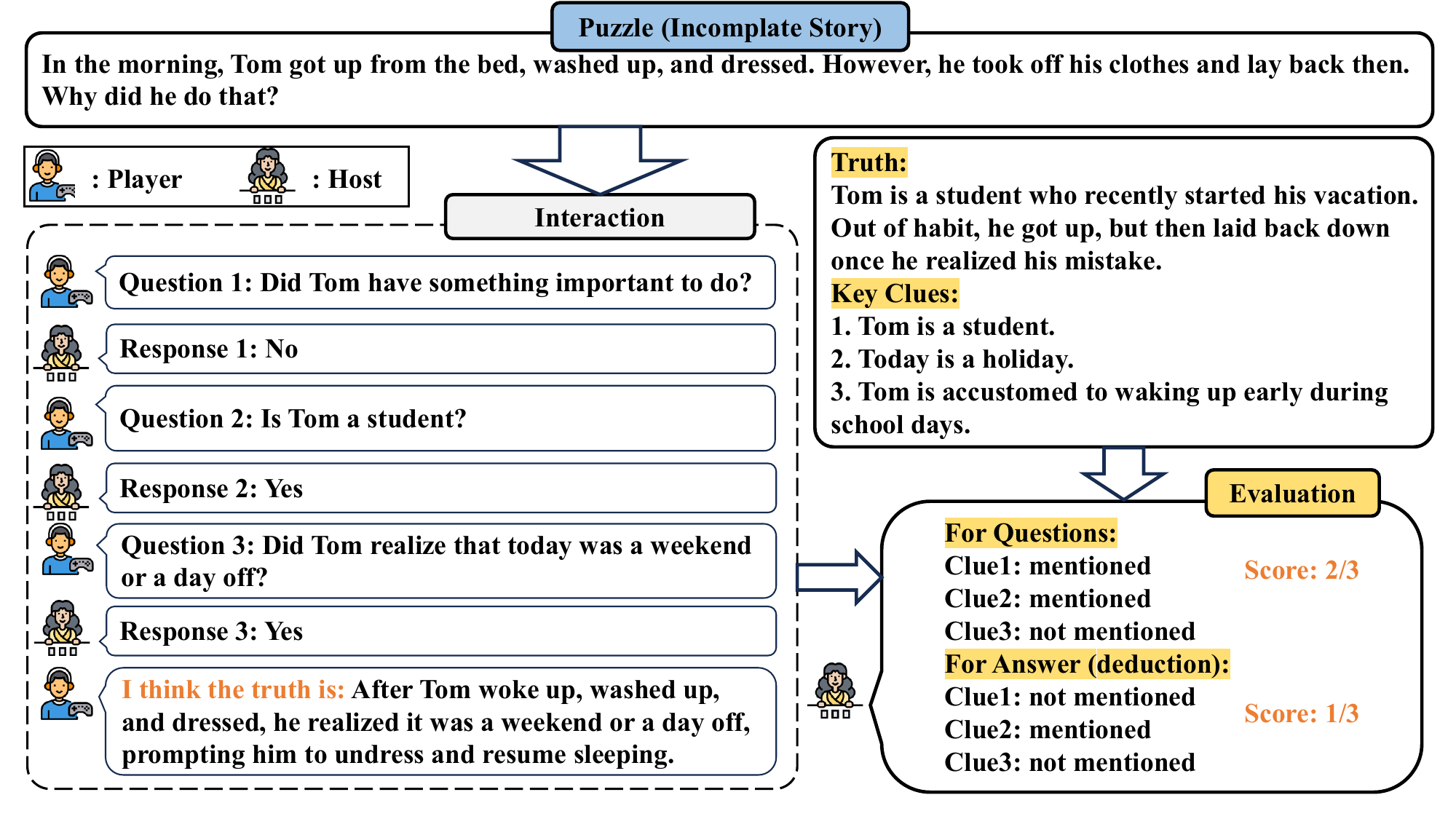}
\caption{An example of a Lateral Thinking Puzzle in our benchmark, including several turns of interaction between the host and the player, and the automatic evaluation for posed questions and player's answer.}
\label{fig:Method_Figure}
\end{figure*}

Specifically, we collect more than 2,000 samples in English and Chinese from diverse Lateral Thinking Puzzles websites, encompassing both incomplete stories (i.e., puzzles) and truths. Combining with LLM and human annotation, we filter original samples: (1) Remove duplicate and irrational samples. (2) Exclude terrifying, bloody, and extreme samples to ensure harmlessness. (3) Test whether LLMs directly generate the truth from the puzzle, and remove samples memorized by LLMs. The final dataset consists of 325 high-quality and challenging samples. Additionally, we annotate key clues for each story to facilitate evaluation.

Within our proposed benchmark, the model under evaluation is expected to demonstrate the following detailed characteristics: 
(1) posing high-quality questions that break out of conventional norms but are beneficial for puzzle-solving.
(2) incorporating available information to progressively arrive at the truth through reasoning.
Therefore, we use the following metrics to evaluate the model: (1) The relevance between the question posed in the multiple turns and the key clues of the truth. (2) The diversity of the question posed in the multiple turns. (3) The consistency between \hsl{its} answer (i.e., deduction) and truth. \hsl{(4) The number of questions posed by the player before making a deduction.}
These \hsl{four} metrics serve to evaluate the model's lateral thinking ability during interaction from various viewpoints, encompassing identifying critical information, the extent of thinking divergence, and integrating information to complete reasoning. 
Figure~\ref{fig:Method_Figure} presents an example process of our benchmark.



Experimental results reveal that the majority of LLMs exhibit almost negligible ability for lateral thinking during interactions. 
Even the most advanced LLM, GPT-4, struggles to deliver good results, and for most open-source models, only completing this task is quite difficult.
This emphasizes \BenchmarkName{}'s challenging feature and its capacity to distinguish the performance of LLMs.
\BenchmarkName{} presents a challenging task for LLMs: how to engage in lateral thinking during interactions, actively pose questions to acquire information, and subsequently uncover the truth.

\section{Task Setups}
Lateral Thinking Puzzle is a popular game among enthusiasts of inference, involving one host and multiple players.
The host is aware of the incomplete story (i.e., puzzle) and truth, while player should pose questions to host based on the puzzle. The host's responses are limited to ``yes'', ``no'' or ``irrelevant'' until player pieces together the truth.

Inspired by the game, in our benchmark, we set one host and one player. 
The model under evaluation acts as the player, while another powerful LLM (e.g., GPT-4, GPT-3.5) serves as the host, engaging in interactions with the model being evaluated.
Specifically, our task is divided into three steps:


(1) \textbf{Rule Introduction}: We use the prompt to introduce basic rules of Lateral Thinking Puzzles and the response format to the player and the host respectively. 
For the player, it is provided with the puzzle and is guided to make a deduction or continue posing the question.
The player is encouraged to pose questions to the host to obtain unknown information.
For the host, it is informed of both the puzzle and the truth. It is also guided to respond appropriately to the player's questions, with responses limited to ``yes'', ``no'' or ``irrelevant'', without actively providing additional information.

(2) \textbf{Player-Host Interaction}: We allow the player and the host to interact for several turns. In each turn, the player poses a question, and the host responds according to the rules. The player acquires unknown information through questioning.

(3) \textbf{Performance Evaluation}: Upon confirming sufficient information acquisition or reaching the maximum interaction rounds, the player is prompted to provide a deduction about the truth, thus finishing the game. At this stage, It is necessary to evaluate the player's performance.


An ideal player not only \hsl{proposes} relevant and divergent questions but also \hsl{integrates} acquired information to make a correct deduction.
Therefore, the following metrics are utilized for evaluation:
(1) \textbf{Question Relevance:} 
The relevance between the questions posed in multiple turns and the truth, which reflects the extent to which posed questions contribute to puzzle-solving.
(2) \textbf{Question Divergence}: The diversity of the question posed in multiple turns, indicating the extent of thinking divergence.
(3) \textbf{Answer Consistency}: The consistency between the player's answer (i.e., deduction) and the truth. 
This metric reflects the player model's ability to gradually reason and integrate available information.
(4) \textbf{Average Turns}: The number of questions that the player proposed before making a deduction.
Too few turns suggest the player model cannot gather sufficient information, while too many turns indicate a stagnation in the model's reasoning process. Therefore, we expect a moderate number of Average Turns.
More details for the above metrics are introduced in Section~\ref{sec:eval}.

\section{\BenchmarkName{} Dataset}
\subsection{Dataset Construction}
We collect more than 2,000 samples from various Lateral Thinking Puzzles websites in English and Chinese.
We review the licenses of the websites to ensure the legality of the data for our non-profit academic research.
Our dataset is rigorously selected following these criteria:

(1) We remove duplicate entries and the entries that contain irrational storylines.

(2) Lateral Thinking Puzzles, favored by inference enthusiasts, usually adopt the characteristics of detective fiction and incorporate horror plots. Some of them are even extremely gory and terrifying. 
To ensure the harmlessness and ethical compliance, as well as to prevent LLMs from refusing to respond, we exclude excessively negative data. 
For dataset diversity, we intentionally select some positive and heartwarming entries, not limiting it to just negative ones.

(3) Considering the extensive coverage of LLMs' training corpus, some classical Lateral Thinking Puzzles are even included in their training data. In order to ensure a fair and authentic assessment, we subject each entry to a model awareness test. 
\hsl{To be specific, we provide LLMs with incomplete stories and directly assessed their responses. Any data where a model displays prior knowledge of the answer is removed from our dataset. 
}

The processed dataset comprises of 325 entries. Furthermore, we annotate each entry in the dataset with key clues, using a combined approach of manual annotation and LLMs assistance. 
These key clues are the key points of the truth, required for deducing the truth from the puzzle.
With these clues, we quantitatively assess the model's performance, similar to a teacher grading a paper based on whether each point is achieved.

\subsection{Dataset Analysis}

We report the statistics of dataset in \BenchmarkName{} in Table~\ref{tab:data_statistics}.
Our dataset includes Chinese and English entries respectively, with data in each language sourced from native content on Chinese and English websites rather than being translations of each other.
Our Chinese and English data capture the cultural characteristics of each language, contributing to the diversity of the dataset.
\begin{table}[ht]
\small
\centering
\begin{tabular}{lcc}
\toprule
& English & Chinese \\ \midrule
\# Entries & 225 & 100\\
Puzzles' Average Length & 29.4 & 42.0 \\
Truths' Average Length  & 51.4 & 66.6  \\
Average Key Clues & 3.7 & 3.8 \\
\bottomrule
\end{tabular}
\caption{Statistics of \BenchmarkName{} dataset.}
\label{tab:data_statistics}
\end{table}

To validate the quality of our dataset, we hire 3 college students to annotate the difficulty of these 325 entries.
The annotation results reveal that only 12\% of the data is easily associated with the truth for human. 
The remaining 88\% is relatively challenging, requiring the vivid imagination and exploration of various possibilities. The annotations underscore the high quality of our dataset, which is well-suited for lateral thinking evaluation.

\section{Experiments}
\subsection{Experiment Setup}
We use entries from our dataset to evaluate lateral thinking ability of mainstream LLMs.
For the host model, we employ GPT-4 or GPT-3.5 because of their superior performance.
The player model is the model under evaluation.
All the evaluated LLMs do not \hsl{undergo} further fine-tuning.
Furthermore, all the player models are guided and introduced to the rules using \hsl{the} same prompt.


\BenchmarkName{} assesses the lateral thinking of LLMs during the interaction, inherently placing the requirement on the model's chat and reasoning capabilities. Consequently, we conduct evaluations on several mainstream chat-based models which act as player models, including:

\textbf{GPT-3.5 and GPT-4}~\cite{OpenAI2023GPT4TR} are two knowledge-rich and advanced models released by OpenAI. 
\textbf{Claude}~\cite{bai2022constitutional} is a series of transformers-based LLMs developed by Anthropic. The models are trained via Constitutional AI to improve helpfulness, honesty and harmlessness.
\textbf{ChatGLM2}~\cite{du2022glm} is a mainstream model that handle both English and Chinese.
We conduct evaluation using ChatGLM2-6B.
\textbf{Llama2-chat}~\cite{touvron2023llama} is a series models that attract significant attention. We conduct evaluation with varying parameter scales: 7B, 13B and 70B.
\textbf{Baichuan-chat}~\cite{baichuan2023baichuan2} is a series of open-source and commercially viable language models.
We conduct experiments on Baichuan-chat-13B and Baichuan2-chat-13B.
\textbf{InternLM-chat}~\cite{2023internlm} is a series of lightweight open-sourced pretrained models without extensive dependencies.
We conduct evaluation of InternLM-chat-20B.
\textbf{Bloomchat-176B}~\cite{bloomchat} is a multilingual chat model based on BLOOM.

We utilize the official API to employ closed-source models, while open-source models are downloaded and run on 1-4 Nvidia A100 GPUs (80GB).
When utilizing LLMs to act as the host and the player, we employ nucleus sampling during the interaction.
To ensure the accurate generation of the host model, we set the temperature to be 0.3 and top\_p to be 0.7. 
To encourage diversified generation by the player model, we set the temperature to be 0.7 and top\_p to be 0.9.
To prevent the interaction from not ending, the maximum number of interaction turns is set to 20.

\subsection{Evaluation Metrics}\label{sec:eval}
To evaluate the lateral thinking ability of the above models, we consider the following metrics:

\noindent\textbf{Answer Consistency (AC)} 
reveals whether the answer provided by the player model is consistent with the truth.
This metric evaluates how many key clues of the truth are included in the player's answer (i.e., deduction). The evaluation is performed by the host model automatically.
\begin{equation*}
    \text{AC} = \frac{1}{|c|} \sum_{i=1}^{|c|}{I(c_i, a)},
\end{equation*}
where $c$ represents the annotated key clues, $a$ represents player's answer, and $I(c_i, a)$ is an indicator function that determines whether $c_i$ is mentioned by $a$. When the host concludes that $c_i$ is mentioned in $a$, $I(c_i, a) = 1$, otherwise $I(c_i, a) = 0$.

\noindent\textbf{Question Relevance (QR)} 
evaluates whether the questions raised by the player are related to the truth.
We utilize the host to assess whether each key clue is related to any raised questions.
\begin{equation*}
    \text{QR} = \frac{1}{|c|}\sum_{i=1}^{|c|} {\mathop{\max}\limits_{j=1}\limits^{|q|}{I(c_i, q_j)}},
\end{equation*}
where $q$ represents all questions posed by the player, and $I(c_i, q_j)$ is an indicator function that determines whether $c_i$ is related to $q_j$.

\noindent\textbf{Question Divergence (QD)} assesses the divergence in the questions posed by the player.
After removing stop-words and punctuation, we compute the pairwise similarity of all the player-posed questions and quantify the divergence of these questions by subtracting the average similarity from 1:
\begin{equation*}
    \text{QD} = 1 - \frac{2}{|q|\cdot(|q|-1)}\sum^{|q|}_{i=1}\sum^{|q|}_{j=i+1} {\text{sim}(q_i, q_j)},
\end{equation*}
where $\text{sim}(q_i, q_j)$ represents the similarity between the two questions $q_i$ and $q_j$. Jaccard Index is employed as the similarity function in our experiments.

\noindent\textbf{Average Turns (AT)} analyzes the average number of interaction turns in Lateral Thinking Puzzles.


\begin{table*}
\small
\centering
\begin{tabular}{clcccccccc} 
\toprule
\multicolumn{1}{l}{\multirow{2}{*}{}} & \multirow{2}{*}{Players} & \multicolumn{4}{c}{English} & \multicolumn{4}{c}{Chinese} \\ 
\cmidrule(l){3-10}
\multicolumn{1}{l}{} &  & AC & QR & QD & AT & AC & QR & QD & AT \\ 
\midrule
\multirow{3}{*}{\begin{tabular}[c]{@{}c@{}}Closed-source\\Models\end{tabular}} & GPT-4 & \textbf{34.2} & \textbf{58.4} & 79.8 & 10.2 & \textbf{36.6} & \textbf{83.8} & \textbf{77.1} & 12.4 \\
 & GPT-3.5 & 17.1 & 53.3 & 77.2 & 13.0 & 9.5 & 67.6 & 76.7 & 10.3 \\
 & Claude & 14.5 & 49.6 & \textbf{81.1} & 8.3 & 16.1 & 54.2 & 68.6 & 7.4 \\ 
\midrule
\multirow{8}{*}{\begin{tabular}[c]{@{}c@{}}Open-source\\Models\end{tabular}} & Llama2-chat-7B* & 11.7 & 30.8 & 74.2 & 15.2 & 8.6 & 66.6 & 62.3 & 15.7 \\
 & Llama2-chat-13B* & 6.7 & \textbf{42.5} & 78.5 & 13.0 & 9.1 & 56.2 & 57.5 & 12.5 \\
 & Llama2-chat-70B* & 11.9 & 38.4 & 73.0 & 14.3 & 10.4 & \textbf{72.6} & 61.2 & 15.2 \\
 & Baichuan-chat-13B & 7.4 & 22.9 & \textbf{91.4} & 8.5 & 4.5 & 12.0 & 53.0 & 9.5 \\
 & Baichuan2-chat-13B & 9.7 & 36.0 & 82.5 & 9.6 & 4.8 & 39.8 & \textbf{72.7} & 8.7 \\
 & InternLM-chat-20B & \textbf{13.7} & 32.0 & 61.1 & 4.7 & \textbf{11.0} & 23.8 & 34.7 & 5.0 \\
 & ChatGLM2-6B & 1.3 & 13.6 & 40.9 & 19.1 & 3.2 & 24.5 & 55.2 & 11.6 \\
 & Bloomchat-176B & 1.8 & 7.9 & 45.7 & 3.4 & 0.0 & 5.0 & 0.0 & 4.1 \\
\bottomrule
\end{tabular}
\caption{Lateral thinking performance of various LLMs with GPT-4 as the host. We report four metrics: AC (Answer Consistency), QR (Question Relevance), QD (Question Divergence), AT (Average Turns). ``*'' indicates that on Chinese datasets, Llama2-chat series, acting as player models, only interact in English with the host model who uses Chinese.}
\label{tab:all gpt4}
\end{table*}

\begin{table*}
\small
\centering
\begin{tabular}{clcccccccc} 
\toprule
\multicolumn{1}{l}{\multirow{2}{*}{}} & \multirow{2}{*}{Players} & \multicolumn{4}{c}{English} & \multicolumn{4}{c}{Chinese} \\ 
\cmidrule(l){3-10}
\multicolumn{1}{l}{} &  & AC & QR & QD & AT & AC & QR & QD & AT \\ 
\midrule
\multirow{3}{*}{\begin{tabular}[c]{@{}c@{}}Closed-source\\Models\end{tabular}} & GPT-4 & \textbf{27.5} & \textbf{72.7} & \textbf{79.6} & 11.7 & \textbf{25.7} & \textbf{70.4} & \textbf{77.9} & 12.8 \\
 & GPT-3.5 & 23.9 & 59.3 & 78.3 & 13.2 & 12.6 & 61.2 & 68.1 & 12.6 \\
 & Claude & 19.6 & 50.2 & 77.9 & 8.4 & 19.6 & 52.2 & 64.7 & 7.2 \\ 
\midrule
\multirow{8}{*}{\begin{tabular}[c]{@{}c@{}}Open-source\\Models\end{tabular}} & Llama2-chat-7B* & 5.7 & 43.2 & 69.6 & 9.5 & 5.7 & 28.0 & 44.2 & 7.9 \\
 & Llama2-chat-13B* & 6.4 & 34.7 & 58.7 & 4.3 & 3.6 & 20.1 & 18.4 & 2.4 \\
 & Llama2-chat-70B* & 13.4 & \textbf{52.2} & 75.5 & 11.0 & 9.9 & \textbf{46.1} & 61.7 & 8.4 \\
 & Baichuan-chat-13B & 6.3 & 20.7 & \textbf{82.3} & 5.4 & 1.5 & 5.5 & 33.4 & 7.0 \\
 & Baichuan2-chat-13B & 11.8 & 45.0 & 77.0 & 9.3 & 10.8 & 42.1 & \textbf{62.0} & 9.4 \\
 & InternLM-chat-20B & \textbf{13.8} & 28.1 & 54.8 & 5.2 & \textbf{13.3} & 27.4 & 34.7 & 16.3 \\
 & ChatGLM2-6B & 1.0 & 3.1 & 55.8 & 12.1 & 5.1 & 28.4 & 41.6 & 8.8 \\
 & Bloomchat-176B & 1.5 & 3.4 & 39.9 & 3.8 & 0.7 & 3.5 & 13.3 & 6.4 \\
\bottomrule
\end{tabular}
\caption{Lateral thinking performance of various LLMs with GPT-3.5 as the host. 
}
\label{tab:all gpt35}
\end{table*}


\subsection{Results Analysis}

We employ GPT-4 and GPT-3.5 as the host models respectively to evaluate the lateral thinking performance of various player LLMs, as shown in Table~\ref{tab:all gpt4} and Table~\ref{tab:all gpt35}.
With GPT-4 and GPT-3.5 as hosts, the overall trends are generally similar.

In terms of Answer Consistency (AC), GPT-4 as the player model exhibits the best performance both in Chinese and English data.
GPT-3.5 and Claude perform similarly, while among the open-source models, InternLM-chat-20B and Llama2-chat-70B exhibit comparable performance. 
The remaining open-source models show poor performance. All other models exhibit a considerable gap with GPT-4.
\hsl{The results suggest that the majority of existing models still lack lateral thinking and are unable to perfectly complete this task.}
We also observe that even with a large parameter, Bloomchat-176B tend to fabricate content when confronted with incomplete information. 
This is also one of the poor-performing models that our proposed benchmark aims to differentiate.

The Question Relevance (QR) score, on the other hand, provide a finer granular reflection of whether each question posed by the model is close to the truth. 
The overall trend of QR generally aligns with the AC, indicating that models proficient in asking questions tend to provide excellent answers.
The QR value is significantly higher than the AC value, indicating that most models lack the ability to extract essential information from questioning and sometimes posing questions for the sake of questioning itself rather than puzzle-solving.
For Claude, there are occasional instances where the QR value is lower than that of GPT-3.5, while the AC value is relatively higher. We observe that when Claude provides answers, it tends to summarize the questions posed before and integrates the useful information.
For Claude, being accustomed to summarizing existing information in the answer makes it easier to cover key clues. This explain why Claude has higher AC values in some cases.

\begin{table*}
\small
\centering
\begin{tabular}{lcccccc} 
\toprule
\multirow{2}{*}{Models} & \multicolumn{2}{c}{Answer Consistency} & \multicolumn{2}{c}{Question Relevance} & \multicolumn{2}{c}{Host Correction Accuracy}\\ 
 \cmidrule(r){2-3} \cmidrule(r){4-5}  \cmidrule(r){6-7} 
 & Score & Correlation & Score & Correlation & GPT-4 & GPT-3.5 \\ 
\midrule
GPT-4 & 43.9 & 0.84$\dag$ & 67.0 & 0.80$\dag$ & 98.4 & 89.8 \\
GPT-3.5 & 9.7 & 0.71$\dag$ & 46.3 & 0.77$\dag$ & 96.8 & 94.5\\ 
\midrule
Baichuan2-chat-13B & 7.0 & 0.69$\dag$ & 24.3 & 0.81$\dag$ & 94.6 & 86.8\\
Baichuan-chat-13B & 5.7 & 0.69$\dag$ & 15.3 & 0.87$\dag$ & 90.4 & 84.5 \\
Llama2-chat-70B & 12.0 & 0.71$\dag$ & 45.5 & 0.81$\dag$ & 94.7 & 89.0 \\
Bloomchat-176B  & 0.0 & - & 5.4 & 0.76$\dag$ & 90.1 & 81.0\\
\bottomrule
\end{tabular}
\caption{Human evaluation results. We employ Spearman Correlation Coefficient to measure the correlation between human evaluation score and GPT-4 automatic evaluation score. $\dag$ represents $p < 0.01$ in significance test. For Host Correction Accuracy, we annotate the accuracy of responses when GPT-4 and GPT-3.5 serve as the host respectively.}
\label{tab:human}
\end{table*}

\begin{figure*}
\centering
\subfigure[ROUGE -- Clues] 
{ 
\includegraphics[width=0.43\linewidth]{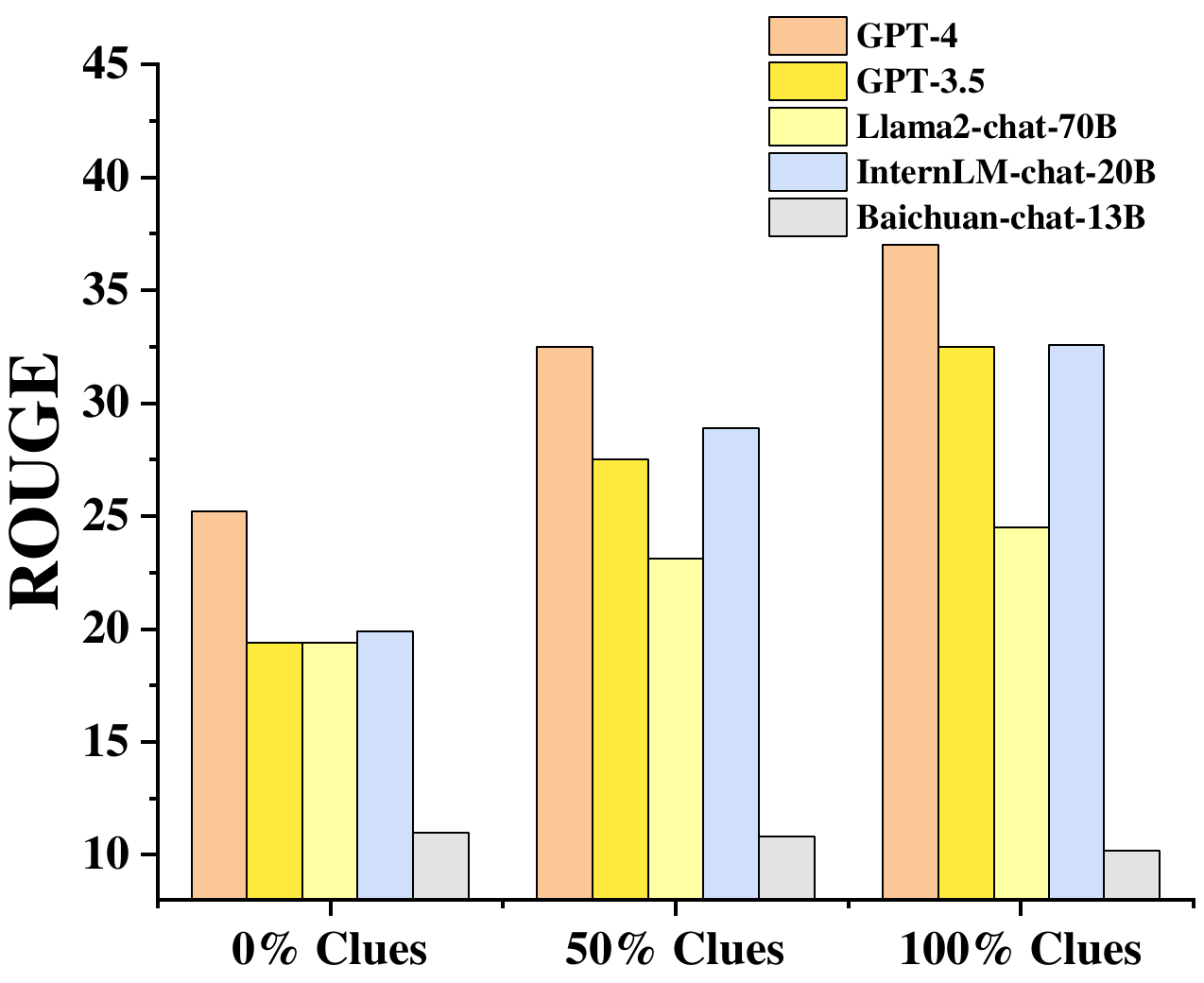} 
} 
\subfigure[Average Turns -- Clues] 
{ 
\includegraphics[width=0.43\linewidth]{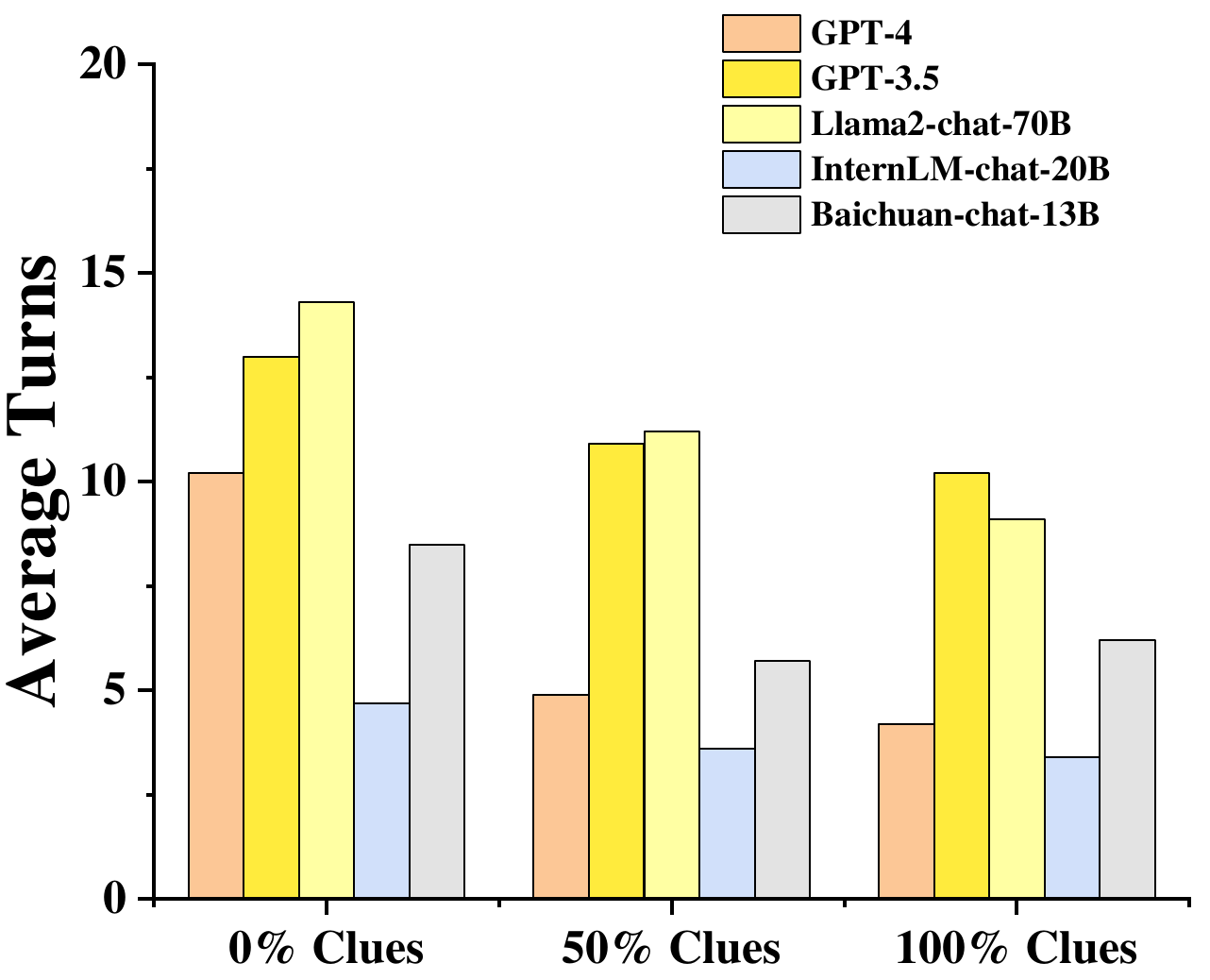} 
} 
\caption{Lateral thinking performance of various LLMs under various difficulty settings: providing 0\% Clues, 50\% Clues and 100\% Clues. We report two metrics: ROUGE and Average Turns.} 
\label{fig:clues} 
\end{figure*}

Question Divergence (QD) score reflects the diversity of questions posed in each turn. 
To solve Lateral Thinking Puzzles, models are encouraged to explore multiple perspectives rather than focusing on a single aspect. Results show that most models exhibit the high QD, indicating their tendency to consider various possibilities. This suggests that they have substantial potential for effectively addressing Lateral Thinking Puzzles.


Average Turns (AT) reflects the number of questions posed by the model. We consider that a model well-suited for solving Lateral Thinking Puzzles should achieve a balanced AT. 
A low AT signifies the model's lack of attempts to acquire information through questioning, while a high AT indicates that the model's questioning fails to hit the crucial aspects of the truth.
For instance, in English data, Bloomchat-176B and ChatGLM2-6B posed the fewest and the most number of questions, respectively. However, the answer generated by them are far from matching the truth.

Furthermore, we conduct experiments on both English and Chinese data which incorporate each local characteristics.
We observe that most models exhibit similar performance on Chinese and English data.
Note that on Chinese data, Llama2-chat series, serving as player models, only communicate in English with the host model who uses Chinese. 
Interestingly, they are capable of effectively conveying meaning to each other, even when using different languages.




\begin{table*}
\small
\centering
\begin{tabular}{l|l} 
\toprule
\multicolumn{2}{l}{\begin{tabular}[c]{@{}p{0.96\linewidth}@{}}\textbf{Puzzle}: A riverboat filled with passengers suddenly capsized, drowning most of those aboard. Why?\\\textbf{Truth}: The boat was moving along a river when a large snake dropped onto the deck. The passengers all rushed to the other side of the boat, thereby overturning it.\\\textbf{Key Clues}: 1. A large snake dropped onto the deck. 2. Passengers rushed to the other side of the boat in panic. 3. The boat capsized due to the sudden shift in weight.\end{tabular}} \\ 
\midrule
\begin{tabular}[c]{@{}p{0.46\linewidth}@{}}
\sethlcolor{yellow}\hl{\textbf{GPT-4:}}\\
\textbf{Host:} \textbf{\{Puzzle\}~\{Rules Introduction\}} Please begin your questions.\\\textbf{Player:} Question 1: Was the boat overloaded with passengers?\\\textbf{Host:} Response 1: No\\\textbf{Player:} Question 2: Was there a storm or rough weather that caused the boat to capsize?\\\textbf{Host:} Response 2: No\\\textbf{Player:} Question 3: Did the boat capsize due to a collision with another object or vessel?\\\textbf{Host:} Response 3: No\\\textbf{Player:} Question 4: Was there a mechanical or structural failure in the boat that caused it to capsize?\\\textbf{Host:} Response 4: No\\\textbf{Player:} Question 5: Did the passengers suddenly move to one side of the boat causing it to capsize?\\\textbf{Host:} Response 5: Yes\\\textbf{Player:} \ul{Answer: The riverboat capsized because all or most passengers suddenly moved to one side of the boat, causing it to lose balance and capsize.}\end{tabular} & \begin{tabular}[c]{@{}p{0.46\linewidth}@{}}
\sethlcolor{yellow}\hl{\textbf{GPT-3.5:}}\\
\textbf{Host:} \textbf{\{Puzzle\}~\{Rules Introduction\}} Please begin your questions.\\\textbf{Player:} Question 1: Did the riverboat capsize due to a natural disaster such as a storm or earthquake?\\\textbf{Host:} Response 1: No\\
...\\
\textbf{Player:} Question 7: Did something unexpected occur during the voyage which caused the riverboat to capsize?\\\textbf{Host:} Response 7: Yes\\\textbf{Player:} Question 8: Was there an animal involved in the incident?\\\textbf{Host:} Response 8: Yes\\\textbf{Player:} Question 9: Did the animal attack the passengers and cause the riverboat to capsize?\\\textbf{Host:} Response 9: No\\\textbf{Player:} Question 10: Did the animal accidentally cause the riverboat to capsize?\\\textbf{Host:} Response 10: Yes\\\textbf{Player:} \ul{Answer: An animal accidentally caused the riverboat to capsize.}\end{tabular} \\ 
\midrule
\begin{tabular}[c]{@{}p{0.46\linewidth}@{}}
\sethlcolor{yellow}\hl{\textbf{Llama2-chat-70B:}}\\
\textbf{Host:} \textbf{\{Puzzle\}~\{Rules Introduction\}} Please begin your questions.\\\textbf{Player:} Sure, I'd be happy to play a game of Lateral Thinking Puzzle with you! Here's my first question: Question 1: Is the reason for the riverboat capsizing related to the boat itself?\\\textbf{Host:} Response 1: No\\\textbf{Player:} Great, that helps to narrow it down. Here's my next question: Question 2: Is the reason for the riverboat capsizing related to the passengers?\\\textbf{Host:} Response 2: Yes\\\textbf{Player:} Great, that helps to narrow it down even further. Here's my next question: Question 3: Did the passengers do something to cause the boat to capsize?\\\textbf{Host:} Response 3: Yes\\\textbf{Player:} Great, now I think I know the answer. Here's my answer: \ul{Answer: The passengers stood up and rushed to one side of the boat, causing it to capsize. Is that correct?}\end{tabular} & \begin{tabular}[c]{@{}p{0.46\linewidth}@{}}
\sethlcolor{yellow}\hl{\textbf{Baichuan-chat-13B:}}\\
\textbf{Host:} \textbf{\{Puzzle\}~\{Rules Introduction\}} Please begin your questions.\\\textbf{Player:} Question 1: What happened?\\\textbf{Host:} Response 1: Irrelevant\\\textbf{Player:} Question 2: How many people were there in the boat?\\\textbf{Host:} Response 2: Irrelevant\\\textbf{Player:} Question 3: Were any life jackets worn?\\\textbf{Host:} Response 3: Irrelevant\\\textbf{Player:} Question 4: Was anyone drunk while operating the ship? \\\textbf{Host:} Response 4: No\\\textbf{Player:} Question 5: Did someone accidentally hit something beneath the water without realizing it? \\\textbf{Host:} Response 5: No\\\textbf{Player:} \ul{Answer: The reason why the passenger steamboat overturned was because some object fell into the hull below the surface and cracked the bottom of the vessel, causing rapid flooding and capsizing.}\end{tabular} \\
\bottomrule
\end{tabular}
\caption{The interaction case between the player and the host in \BenchmarkName{}. We present GPT-4, GPT-3.5, Llama2-chat-70B, and Baichuan-chat-13B as the respective player models in the case.}
\label{table:samples}
\end{table*}

\subsection{Human Evaluation}

We design the human evaluation to assess the lateral thinking ability of player models and to validate the appropriateness of the host model both in responses and evaluation.
Specifically, we select the interactions of 50 entries between various models and host model during puzzle-solving, from Chinese datasets. 
We provide a detailed explanation of task rules and the standards for annotating metrics to 3 undergraduate students, and hire them for manual annotation.
The annotation encompasses: 
(1) the relevance of the player model's questions to the truth's key clues,
(2) the consistency between the player model's answer and truth's key clues,
(3) the accuracy of the host model's responses to the player model's questions. 
Notably, (1) and (2) are formally consistent with QR and AC metrics for automated evaluation by the host model, respectively. We calculate the Spearman correlation coefficient between the results of manual evaluation and GPT-4's automated evaluation.

Table~\ref{tab:human} demonstrates a strong correlation between the human evaluation and the GPT-4 automated evaluation. 
Additionally, the relative rankings among different models are generally consistent, indicating the rationality of AC and QR metric and the validity of employing GPT-4 for the automated evaluation. 
Furthermore, GPT-4 achieves an accuracy rate of over 90\% as a host responding to questions, underscoring its competence as an effective host for Lateral Thinking Puzzles.
Meanwhile, GPT-3.5 achieves an accuracy rate of over 80\% as a host responding to questions, lower than GPT-4. Although the responses from GPT-3.5 are generally correct, there are some instances of misleading the players. 
Note that when faced with players like Bloomchat and Baichuan-chat-13B, which lack lateral thinking ability, the host model tends to make more errors in responses.

\subsection{Fine-grained Analysis}

Most models still lack proficient Lateral thinking, making it challenging to directly accomplish Lateral Thinking Puzzle. To further refine model evaluations, we assess the model's lateral thinking ability at varying levels of difficulty by introducing settings with different numbers of clues.

Specifically, in the initial phase of interaction, we present the puzzle alongside either 50\% or 100\% of the clues, comparing their performance to a setting with 0\% clues. The previous introduced evaluations shown in Table~\ref{tab:all gpt4} and Table~\ref{tab:all gpt35} are conducted under 0\% clues setting.
In this graded evaluation, we calculate ROUGE-L~\cite{lin-2004-rouge} between deduction of the player model and the truth, utilizing English data and GPT-4 as the host model.

Figure~\ref{fig:clues} shows the performance of different models under various difficulty settings.
We observe that for models such as GPT-4 and GPT-3.5, as the amount of provided clues increases, the deduction becomes closer to the truth. This trend reflects the strong information integration capability of these models. 
However, for Baichuan-chat-13B, even when provided with all clues, meaning the complete information, it still cannot deduce the truth from the provided information. The performance under 100\% clues setting represents the upper limit of each model's performance in Lateral Thinking Puzzle.
It indicates that Baichuan-chat-13B lacks the ability to comprehend complicated instructions and integrate the available information.

Furthermore, with the increase in the number of provided clues, models that are skilled at integrating information exhibit the decrease in Average Turns. It's because the more clues you provide, the less you need to ask for information. Therefore, for powerful models, more clues settings result in fewer Average Turns.



\subsection{Case Study}
Table~\ref{table:samples} shows the case of different player models tackling Lateral Thinking Puzzle, including the puzzle, truth, key clues and interactions between host model and player model.
The case demonstrates that there is a substantial gap between the puzzle and the truth, including a few pieces of key information. 
In order to get the truth, the player needs to think divergently, explore various possibilities, and strategically structure its questions to acquire crucial information. It's evident that this task is quite challenging.
During the interaction with the host, player model GPT-4 shows divergent thinking by posing some thought-provoking questions, such as ``Did the passengers suddenly move to one side of the boat causing it to capsize?''. Furthermore, player GPT-4 effectively covers a wide range of aspects by addressing only five questions, including weather-related factors, accidental causes, human-related factors. GPT-4 also references some key clues in the final answer, shows its proficiency in divergent thinking and information integration.

GPT-3.5 also poses some divergent and valuable questions, such as ``Did the animal accidentally cause the riverboat to capsize?''. In the answer, GPT-3.5 covers ``animal'' which is acquired through questioning during the interaction. Llama2-chat-70B tends to ask questions with a narrow focus, and the question format does not strictly adhere to the requirements specified in the prompt, incorporating some unnecessary discourse. Although the answer of Llama2-chat-70B does not confirm to the format requirements, it covers ``rushed to one side of the boat'', which is acquired through questioning.
Baichuan-chat-13B disregards the prompt's repeated emphasis on asking only yes-or-no questions and, without obtaining sufficient information, starts to fabricate the answer.
The above results reveal differences in lateral thinking abilities among various player models. Furthermore, there is still substantial room for improvement in the lateral thinking abilities of existing LLMs.

\section{Related Work}
\noindent\textbf{LLMs Evaluation.}
With the emergence of ChatGPT and GPT-4~\cite{OpenAI2023GPT4TR}, the capabilities of LLMs have become increasingly impressive, drawing significant attention to evaluation of LLMs~\cite{DBLP:journals/corr/abs-2308-10529, DBLP:journals/corr/abs-2308-06966, DBLP:conf/emnlp/YeLZLM0023, li2024rethinking}. 
Recently, many research efforts focuses on evaluating LLMs from various perspectives, including language tasks~\cite{cheng2023mrrl, huang-etal-2022-towards, DBLP:conf/acl/LiZLLLSWLCZ22, ma-etal-2022-linguistic, bang2023multitask}, reasoning~\cite{bang2023multitask,bian2023chatgpt}, robustness~\cite{li2023survey, cheng2023accelerating}, trustworthiness~\cite{hagendorff2023human, cheng2023m}, medical applications~\cite{chervenak2023promise,cascella2023evaluating}, and ethical considerations~\cite{Cao2023AssessingCA,Parrish2021BBQAH,DBLP:journals/patterns/LiuLTLZ22}. 
Currently, mainstream evaluation benchmarks, such as MMLU~\cite{hendrycks2020measuring}and C-Eval~\cite{huang2023c}, assess LLMs from a multi-task perspective, yet are still constrained within a question-answering framework.
However, these benchmarks 
lack evaluation criteria for measuring LLMs' lateral thinking capabilities.

\noindent\textbf{LLMs Reasoning.}
\hsl{In terms of reasoning, some researchers try to guide LLMs' reasoning through Prompt Engineering frameworks like CoT~\cite{wei2022chain}, ToT~\cite{yao2024tree}, GoT~\cite{besta2023graph}, AoT~\cite{sel2023algorithm} and so on. 
The Chain of Thoughts (CoT) sequences LLM decision-making, while the Tree of Thoughts (ToT) simplifies problems into sub-tasks for structured reasoning. The Graph of Thoughts (GoT) maps decisions onto a Directed Acyclic Graph for complex problem-solving, and the Algorithm of Thoughts (AoT) enables LLMs to follow algorithmic reasoning paths for in-context learning.
Additionally, researchers further optimize the capabilities of LLMs Reasoning, for instance, by introducing Chain-of-Thought Self-Consistency~\cite{wang2022self}, Skeleton-of-Thought~\cite{ning2023skeleton}, and Program-of-Thoughts~\cite{chen2022program}.
}

\noindent\textbf{Lateral Thinking.}
\citet{Syahrin2019CreativeTP} propose lateral thinking as a cognitive activity employed to construct creative ideas. Similarly, \citet{Hidayat2018TheEO} indicate that lateral thinking is associated with generating novel ideas, which serves as progressive elements across various scientific domains, from engineering to art, from politics to personal well-being.
\citet{Bono1999LateralT} points out that vertical thinking and lateral thinking are mutually complementary, with vertical thinking being selective, and lateral thinking being creative. Lateral thinking offers alternative perspectives to vertical thinking, thereby augmenting its efficacy~\cite{Relaiza2021CognitivePA}.



\section{Conclusion}
In this paper, we propose a novel interactive benchmark to evaluate the lateral thinking capability of LLMs, following the setup of a popular game named Lateral Thinking Puzzle.
We collect data from various Lateral Thinking Puzzle websites and construct an evaluation dataset.
Experimental results indicate that most LLMs perform poorly on lateral thinking capability, which also demonstrates that our benchmark is challenging and has the capacity to distinguish LLMs' performance.
In the future, we will further investigate how to evaluate the lateral thinking ability of foundation models rather than chat models. 
Additionally, there is immense potential for promising research in improving the lateral thinking capability of LLMs.

\section{Ethical Considerations}
In this paper, we propose a novel evaluation benchmark named \BenchmarkName{} for assessing the lateral thinking capability of LLMs.
All the original samples are collected from publicly accessible and legitimate websites, and do not contain any sensitive information. We check the licenses of the samples to ensure its usability for non-profit academic research. During data processing, we remove all bloody, terrifying and extreme stories, ensuring that each story in our benchmark is harmless and ethical.
Thanks to the significant improvement in data annotation efficiency brought by LLMs, part of data annotation and verification is conducted by the authors ourselves. To mitigate the authors' biases, we hire 3 college students for data annotation during the dataset quality verification and human evaluation stages. The average hourly salary paid to them is approximately twice the local minimum salary.

\hsl{
\section{Acknowledgments}
This research is supported by National Natural Science Foundation of China (Grant No.62276154), Research  Center for Computer Network (Shenzhen) Ministry of Education, the Natural Science Foundation of Guangdong Province (Grant No. 2023A1515012914), Basic Research Fund of Shenzhen City (Grant No. JCYJ20210324120012033 and JSGG20210802154402007), Shenzhen Science and Technology Program (Grant No. WDZC20231128091437002), the Major Key Project of PCL for Experiments and Applications (PCL2021A06), and Overseas Cooperation Research Fund of Tsinghua Shenzhen International Graduate School (HW2021008).}
\section{Bibliographical References}\label{sec:reference}

\bibliographystyle{lrec-coling2024-natbib}
\bibliography{lrec-coling2024-example}

\end{document}